\documentclass{INTERSPEECH2023}

\usepackage{caption}
\usepackage{xcolor}
\usepackage{booktabs}
\usepackage{multirow}
\usepackage{graphicx}
\usepackage{subcaption}
\usepackage{hyperref}

\usepackage{footnote}
\usepackage{footmisc}

\newcommand\blfootnote[1]{%
  \begingroup
  \renewcommand\thefootnote{}\footnote{#1}%
  \addtocounter{footnote}{-1}%
  \endgroup
}

\interspeechcameraready

\title{ProsAudit, a prosodic benchmark for self-supervised speech models}
\name{Maureen de Seyssel$^{1,2}$,
Marvin Lavechin$^{1,5}$,
Hadrien Titeux$^{1}$,
Arthur Thomas$^{\footnotesize{\dag}}$,
Gwendal Virlet$^{4\footnotesize{\dag}}$,
Andrea Santos Revilla$^{\footnotesize{\dag}}$,
Guillaume Wisniewski$^{2}$,
Bogdan Ludusan$^{3}$,
Emmanuel Dupoux$^{1,5}$
}

\address{
   $^1$ Cognitive Machine Learning, ENS–CNRS–EHESS–INRIA–PSL Research University, France\\
   $^2$ Université Paris Cité, CNRS, Laboratoire de linguistique formelle,  Paris, France\\
   $^3$ Faculty of Linguistics and Literary Studies \& CITEC, Bielefeld University, Germany \\
   $^4$ PEGASE, INRAE,  Institut Agro, Saint-Gilles, France \quad
   $^5$ Meta AI Research, France
}

 \email{maureen.deseyssel@gmail.com}

\begin{document}

\maketitle

\begin{abstract}
We present ProsAudit, a benchmark in English to assess structural prosodic knowledge in self-supervised learning (SSL) speech models.  It consists of two subtasks, their corresponding metrics, and an evaluation dataset. In the protosyntax task, the model must correctly identify strong versus weak prosodic boundaries. In the lexical task, the model needs to correctly distinguish between pauses inserted between words and within words. We also provide human evaluation scores on this benchmark. We evaluated a series of SSL models and found that they were all able to perform above chance on both tasks, even when evaluated on an unseen language. However, non-native models performed significantly worse than native ones on the lexical task, highlighting the importance of lexical knowledge in this task. We also found a clear effect of size with models trained on more data performing better in the two subtasks.
\end{abstract}

\section{Introduction}

	\ifinterspeechfinal  \blfootnote{$^\dag$Work performed while the authors were employed at CoML} \vspace{-0.6em} \else    \fi

In recent years, self-supervised learning (SSL) speech models such as  Wav2Vec \cite{baevski2020wav2vec}, CPC \cite{oord2018representation,riviere2020unsupervised}, HuBert \cite{hsu2021hubert} have made groundbreaking advancements, while removing the need for labeled data as they use information extracted from the input raw audio itself.  Multiple benchmarks and metrics have been since developed to test (and exhibit) the linguistic knowledge of such models at different levels. For instance, the Zero-Resource speech challenge \cite{dunbar2017zero,nguyen2020zero,dunbar2021zero} offers zero-shot evaluation metrics at the phonetic, lexical, syntactic and semantic levels and SUPERB \cite{yang2021superb} allows downstream evaluation on speech processing tasks, including paralinguistics and speaker-related tasks.

An important aspect of language that has received little to no attention in SSL models is prosody. As a result, we still know little about SSL encoding of structural prosodic knowledge. Yet, through its key components: rhythm, stress, and intonation, prosody plays a significant role in language processing \cite{cutler1997prosody, dahan2015prosody}, interfacing with other linguistic levels (e.g., lexical, syntactic levels), while also carrying paralinguistic information (e.g., emotion) (for a literature review, see \cite{cutler1997prosody} and \cite{dahan2015prosody}). Recently, \cite{kharitonov2022text} proposed to explicitly implement prosodic knowledge into SSL models by forcing prediction of pitch and duration within the models, but only human evaluation on downstream tasks was used as an indirect cure of prosodic encoding in the models. Besides, \cite{lin2023utility} proposed a benchmark to test pragmatics aspects of prosody of the models using downstream tasks. Yet, there is currently no zero-shot metrics which allows systematic evaluation of prosody in such models.

In this paper, we fill this gap by proposing an evaluation benchmark, ProsAudit, which assesses SSL speech models' ability to learn prosodic information at the structural level. By such, we refer to how prosody contributes to the organisation of speech by marking the boundaries between words, phrases, and sentences. Our benchmark, in English, consists of two subtasks. The protosyntax task tests the model's ability to identify strong versus weak prosodic boundaries (e.g., see \cite{ludusan2021effect} for an evaluation of human performance). The lexical task tests the model's ability to distinguish between pauses inserted between and within words. Being a proxy for detecting word boundary versus word internal, this task requires some lexical knowledge. Crucially, we also provide results on these two tasks carried out on human evaluation. 
\ifinterspeechfinal 
We bring our benchmark to the Spoken Modelling track of the Zero-Resource Speech Challenge~\cite{dunbar2021zero} and present some baseline results, which will be integrated into the leaderboard.
\else  We make our benchmark available online, and present some baseline results. 
\fi
Finally, we conduct further analysis on factors like input quantity and nativeness, putting the models in perspective with prosodic learning in humans.

\section{Methods}

\begin{table*}
\caption{Examples of stimuli in the English benchmark, for the ProtoSyntax and Lexical tasks. Numbers in subscript correspond to prosodic break tiers in ToBI format.}
\label{tab:stimuli_examples}
\centering

\begin{tabular}{@{}lll@{}}
\toprule
task        & condition & stimuli \\ \midrule
protosyntax & natural   & \textsubscript{4}\textit{She}\textsubscript{1} \textit{went}\textsubscript{1} \textit{to}\textsubscript{1} \textit{jail}\textsubscript{4}
\textit{\textbf{$<$PAUSE$>$} }
\textit{for}\textsubscript{1}
\textit{what}\textsubscript{1}
\textit{appeared}\textsubscript{1}
\textit{to}\textsubscript{1}
\textit{be}\textsubscript{1}
\textit{a}\textsubscript{1}
\textit{murder}\textsubscript{4}
     \\
protosyntax & unnatural & \textsubscript{4}\textit{She}\textsubscript{1} \textit{went}\textsubscript{1} \textit{to}\textsubscript{1} \textit{jail}\textsubscript{4}
\textit{for}\textsubscript{1}
\textit{what}\textsubscript{1}
\textit{appeared}\textsubscript{1}
\textit{to}\textsubscript{1}
\textit{\textbf{$<$PAUSE$>$} }
\textit{be}\textsubscript{1}
\textit{a}\textsubscript{1}
\textit{murder}\textsubscript{4}
     \\
lexical     & natural   & \textsubscript{4}\textit{She}\textsubscript{1} \textit{went}\textsubscript{1} \textit{\textbf{$<$PAUSE$>$} } \textit{to}\textsubscript{1} \textit{jail}\textsubscript{4}
\textit{for}\textsubscript{1}
\textit{what}\textsubscript{1}
\textit{appeared}\textsubscript{1}
\textit{to}\textsubscript{1}
\textit{be}\textsubscript{1}
\textit{a}\textsubscript{1}
\textit{murder}\textsubscript{4}      \\
lexical     & unnatural & \textsubscript{4}\textit{She}\textsubscript{1} \textit{went}\textsubscript{1} \textit{to}\textsubscript{1} \textit{jail}\textsubscript{4}
\textit{for}\textsubscript{1}
\textit{what}\textsubscript{1}
\textit{a-}\textit{\textbf{$<$PAUSE$>$} }\textit{-ppeared}\textsubscript{1}
\textit{to}\textsubscript{1}
\textit{be}\textsubscript{1}
\textit{a}\textsubscript{1}
\textit{murder}\textsubscript{4}       \\ \bottomrule
\end{tabular}
\vspace{-4mm}
\end{table*}

\subsection{ProsAudit benchmark}

We created prosodic benchmarks in English, composed of two tasks which each focus on different aspects of structural prosody: the lexical task and the protosyntax task. These tasks are designed to evaluate the models' understanding of prosody by presenting them with pairs of stimuli that only differ in the placement of a pause.

\ifinterspeechfinal \vspace{-0.6em} \else  \fi

\subsubsection{Material and data preprocessing}
We used the Boston University Radio News Corpus (BU) \cite{ostendorf1995boston} dataset to create the evaluation set, as it includes word and phone-level transcriptions along with prosodic hierarchy annotations based on the American English ToBI system \cite{silverman1992tobi}. The dataset is a collection of professionally-read news stories.

 We selected segments based on the criteria outlined in \cite{ludusan2021effect}.
 To qualify, a segment had to start and end with an intonation phrase (IP) boundary (ToBI level 4) and contain one internal prosodic boundary, which could be an IP boundary or an intermediate boundary (ToBI level 3). Additionally, the qualifying segments had to meet specific criteria:  a minimum and maximum duration of 2 and 5 seconds; the internal prosodic boundary could not be between the two first or two last syllables of the utterances, and a pause should be annotated wherever a prosodic IP was annotated\footnote{This is not systematic as IP can be present without the speaker marking a pause}. The next step consisted in automatically deleting all existent annotated pauses from the stimuli, applying some crossfading (10ms on each side of the pause) to prevent abrupt cut-off or jump in the audio.

\ifinterspeechfinal \vspace{-0.6em} \else  \fi

\subsubsection{Creating the protosyntax and lexical tasks}

 In the protosyntax task, one stimulus has a pause placed at a ``natural" location : a prosodic phrase boundary  (ToBI levels 3 or 4). In contrast, the other stimulus has the pause placed at an ``unnatural" location where there are no higher level prosodic breaks (ToBI levels 1 or 2). This task aims to assess the models' understanding of structural prosody at the sentence level.
In the lexical task, the prosodic boundary is present at a word boundary in the natural condition (levels 1 or 2)  and within a word in the unnatural condition. Because the differences between these two conditions are less marked prosodically, lexical knowledge should be required on top of prosodic knowledge to perform well in this task. Examples of both tasks are presented Table \ref{tab:stimuli_examples}.

While there is only one possibility for where the pause can be in the natural condition, multiple stimuli can be created for each pair in the unnatural condition.
Therefore, we included constraints on the position of the break from the start and end of the audio regarding the number of syllables and seconds. We sampled the final stimuli by implementing similarity losses for these constraints, aiming for the smallest divergence between the distributions of the natural and unnatural conditions.
Once the stimuli pairs and break positions were defined, we inserted a 400ms pause with crossfading at the chosen position, resulting in a total of 5,234 pairs in the protosyntax task and 5,178 pairs in the lexical task. Since the pause is artificially inserted in both unnatural and natural conditions, the duration of the two stimuli is the same in both conditions, with the only difference being the location of the pause within the stimuli.
Finally, we randomly sampled about 10\% of the pairs to create a dev set, resulting in 2,355 \textit{(262)} pairs in the protosyntax task for the test \textit{(dev)} set and 2,330 \textit{(259)} pairs in the lexical task for the test \textit{(dev)} set.

\ifinterspeechfinal \vspace{-0.7em} \else  \fi

\subsubsection{Metric}

We employed, in both subtasks, the same metric as the sWuggy and sBlimp metrics proposed in \cite{nguyen2020zero} : we compute, for a pair of stimuli,  the probability of the model evaluated (or pseudo-probability) to generate these stimuli. If the probability of the “natural” stimuli is higher than the unnatural stimuli, we give a score of 1 for this pair, otherwise a score of 0. The final score corresponds to the average score for all dev or test pairs in the given subtask (see \cite{nguyen2020zero} for more details).

\ifinterspeechfinal \vspace{-0.7em} \else  \fi

\subsection{Baselines}

We evaluated several self-supervised learning models of speech on the protosyntax and lexical tasks, as well as a human baseline.

\ifinterspeechfinal \vspace{-0.6em} \else  \fi

\subsubsection{GSLM and pGSLM baselines}

We evaluated our models against the prosody-aware models presented in \cite{kharitonov2022text}, which we refer to as pGSLM models. These models are an extension of the “GSLM” models, presented in an earlier study \cite{lakhotia2021generative}, and that we also evaluated. All of these models are trained on a “clean” 6k hours sub-sample of the Libri-Light dataset \cite{kahn2020libri}, which is made of English audiobooks. The GSLM models have three main components: an acoustic model (HuBert), a quantizer, and a language model (transformer). Here, we present two versions of these GSLM models, the standard version and the “deduplicated” version, where the units output by the quantizer are deduplicated. 
The pGSLM models build upon the GSLM models by adding tasks for the language model, which predicts the fundamental frequency (f0) and duration of the following frames, in addition to the frame's content. Because the duration is predicted, the information is also removed from the deduplicated units.

The f0 and duration prediction tasks should enable the model to learn the rhythm and intonation of the speech, which are important cues for understanding the meaning of speech. Hence, this modification allows the pGSLM models to incorporate prosodic information into their predictions. By incorporating this information, the pGSLM models may be able to produce more natural-sounding speech and make more accurate predictions. In this paper, we consider four versions of these pGSLM models, trained on continuous (cont.) or discrete (disc.) input, and with or without a prosodic shift in the prediction (see \cite{kharitonov2022text} for more details). However, we only compute the pseudo-probability based on the original LM token prediction, similarly to what is done in \cite{kharitonov2022text}.

\ifinterspeechfinal \vspace{-0.6em} \else  \fi

\subsubsection{STELA baselines}

The tasks were evaluated on English and French models from \cite{lavechin2022statistical}, which we refer to as STELA models. By also using models trained in an unseen language (French), we can test the effect of nativeness on structural prosody. These models, trained on English and French audiobooks, use a Contrastive Predictive Coding objective \cite{oord2018representation,riviere2020unsupervised} for the first acoustic model, which generates continuous representations. These representations are then quantized using k-means, and a language model (LSTM) is trained on the resulting units. Additionally, these models are trained on varying amounts of data, allowing for the creation of training size curves\footnote{To ensure comparable results, models at all train sizes are overall trained on the same amount of data, with more models being trained on the smallest train sizes. See \cite{lavechin2022statistical} for more details.}. To ensure a fair comparison with other models, a ``deduplicated" version of the 3,200 hour English model was also trained.

\ifinterspeechfinal \vspace{-0.6em} \else  \fi

\subsubsection{Human evaluation}

Finally, we ran a human evaluation by presenting the same protosyntax and lexical tasks in an online experiment. We used Mechanical Turk to recruit 389 participants for 790 sessions, which sessions were compensated \$1.5 USD each.
In a session, the participant was presented with a series of 7 pairs from the protosyntax and 7 pairs from the lexical task in a random order, along with two examples at the beginning of the session, and four additional 4 controls\footnote{For a control pair, the pause is inserted at an intonation phrase boundary (natural) and within a word (unnatural).} specifically easier in order to filter out participants who were not paying attention to the task. For each pair, a webpage was presented to them, asking them to listen to two audios and decide which of the two audio was the most natural (the condition order was randomised), as well as how sure they were of their decision (slightly, moderately, strongly). We also included two examples at the beginning of the test, as well as 4 controls in order to filter out participants who were not doing the task correctly. Participants were paid \$1.5 USD per session.
For the analyses, we discarded all non-native English participants, as well as all sessions where the participant did not correctly pass at least 5 out of the 6 examples + controls, resulting in a total of 255 participants (515 sessions).
Finally, we only included in our analysis stimuli pairs that were listened to at least 5 times. We refer to these final stimuli as the “human subset”, independent from the dev and test sets, which is composed of 521 pairs for the protosyntax task and one of 510 pairs for the lexical task.

\ifinterspeechfinal \vspace{-0.5em} \else  \fi

\section{Results}

\ifinterspeechfinal \vspace{-0.5em} \else  \fi

\subsection{Benchmark}

\begin{table}[]
\caption{ProsAudit accuracy scores (\%) for the different models.}
\label{tab:benchmark}
\centering
\resizebox{\columnwidth}{!}{
\begin{tabular}{@{}llllll@{}}
\toprule
                           &                 & \multicolumn{2}{l}{protosyntax} & \multicolumn{2}{l}{lexical} \\
model                      & dataset         & dev            & test           & dev          & test         \\ \midrule
STELA       &   3200h audiob. &      \textbf{72.5}     &  \textbf{74.9}     &  68.7      &     68.3         \\
STELA deduplicated            & 3200h audiob. &      58.0        &    58.5            &      48.7  &   46.7          \\

GSLM            & Librilight 6k   &    58.8       &   58.1        &    53.3    & 54.1      \\
GSLM  deduplicated             & Librilight 6k   &     67.2       &66.5         &   73.8     &    70.5    \\
pGSLM - cont.  & Librilight 6k   &    65.7      &  66.8         &  73.8     &    71.5   \\
pGSLM - cont. + shift   & Librilight 6k   &   69.1      &   66.8      &  \textbf{74.9}    &   71.9     \\
pGSLM - disc.  & Librilight 6k   &    67.6      &   64.8        &    74.5     &  71.1    \\
pGSLM - disc. + shift    & Librilight 6k   &   69.1    &   65.9     & 72.6     & \textbf{72.9}    \\ \bottomrule
\end{tabular}
}
\vspace{-5.5mm}
\end{table}

\ifinterspeechfinal 
The benchmark (with both the dev and test sets) is available as part of a new evaluation metric for Track 4 of the Zero-Resource Speech challenge\footnote{\href{https://download.zerospeech.com/datasets/prosaudit-dataset.zip}{https://download.zerospeech.com/datasets/prosaudit-dataset.zip}}, and a new leaderboard is also setup.
\else  
The benchmark (with both the dev and test sets) is available online\footnote{More information along with url to the benchmark will be made available after the double-blind review.}.
\fi

Scores on the protosyntax and lexical prosodic tasks evaluated on English models are presented in Table \ref{tab:benchmark}.
First, all models perform above chance in both the protosyntax and lexical tasks, suggesting that all of these models have some prosodic knowledge about the structure of sentences and words.

An interesting observation is that while the GSLM and pGSLM models perform better on the lexical task, suggesting a better understanding of word boundaries, the STELA models actually perform better on the protosyntax task, indicating a stronger sensitivity to the prosodic structure of sentences. This is surprising given that the pGSLM models are specifically trained to also predict the duration and pitch of the next frames, in addition to the content of the frame. However, it must be reminded here that the pseudo-probability for the pGSLM model is only computed at the token level. A potential improvement for future work would be considering the duration and pitch losses when computing the pseudo-probability.

There is little difference in performance between the four pGSLM models, although the continuous pGSLM with prosodic shift performs slightly better overall. More surprisingly, the GSLM model with deduplicated units performs similarly to the pGSLM models, even though the deduplication process removes the duration information, which is an essential aspect of prosodic information (while the GSLM models are also trained on deduplicated units, the duration information is reinjected as a separate loss). This suggests that the lexical and grammatical information present in the units may be sufficient for the model to perform well on the two tasks, even without the prosodic information. It is possible that the model learned to extract the relevant information from the units even with the duration information removed.

Finally, while deduplicating the units in the GSLM model greatly helps in both tasks, the opposite happens when deduplicating the units with the STELA model, which yields much poorer results. While surprising at first, this could be caused by multiple factors that vary between the two model types: the quantity of data (there is nearly double the amount of data in the GSLM models) and the architecture of the language model (an LSTM in the case of the STELA model and a transformer in the GSLM ones). It would be interesting to study further the impact of deduplication on other language evaluation tasks for these two types of models.

\begin{figure*}[]
  \centering
  \includegraphics[trim={0cm 0.18cm 0cm 0cm},clip,width=0.9\linewidth, scale=0.3]{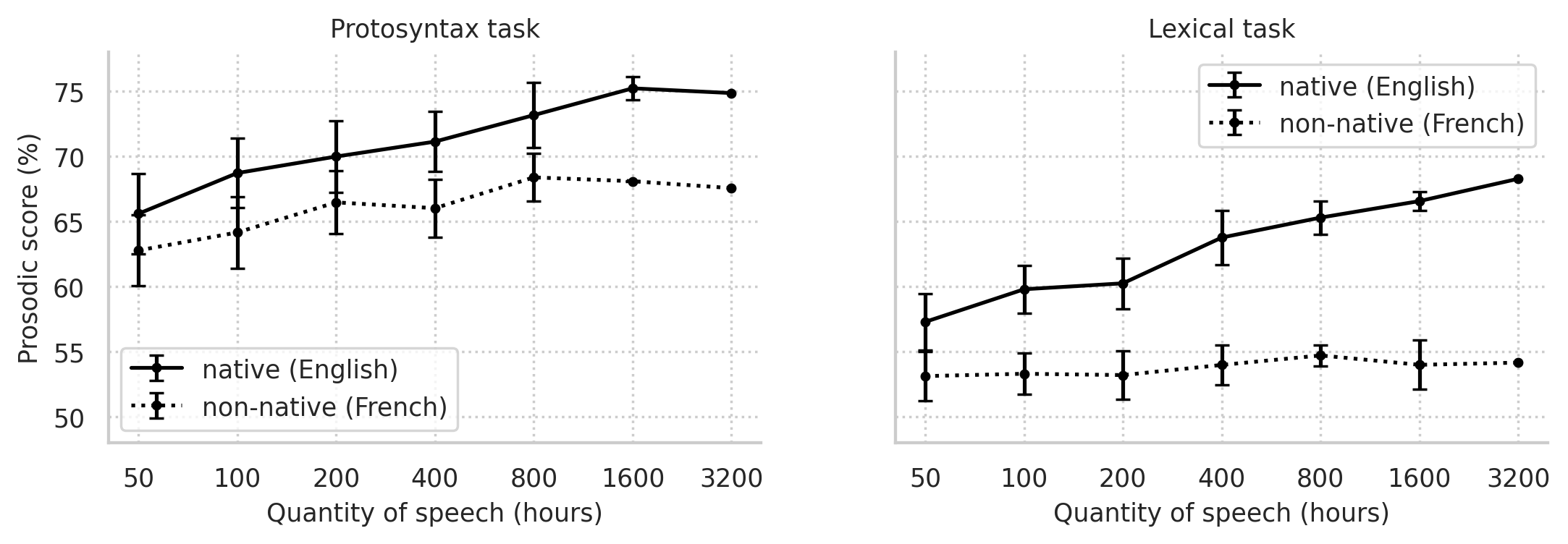}
  \caption{ProsAudit accuracy scores (\%) for the Protosyntax and Lexical subtasks, on the STELA models, w.r.t. the training size.}
  \label{fig:size_effect}
  \vspace{-5mm}
\end{figure*}

\ifinterspeechfinal \vspace{-0.5em} \else  \fi

\subsection{Further analyses}

To better understand the presence of structural prosody in SSL models, we conducted a series of analyses using our ProsAudit benchmark.

\vspace{0.2cm}
\noindent\textbf{Human/Machines comparison.} Table \ref{tab:human_comparison} shows the scores of the human cohort on the lexical and protosyntax tasks, compared to the best-scoring STELA and (p)GSLM models on the same human subset\footnote{Computation of the correlation between human and machine scores for the different items was not possible due to the high variability and insufficient number of responses per items in the human condition.}.  Humans perform better than chance on both tasks. The protosyntax results (80.5\%) are slightly lower than those presented in \cite{ludusan2021effect} (93.2\%), where participants were presented with a Japanese version of the task. However, this can be explained by the fact that the stimuli in \cite{ludusan2021effect} were much more curated and manually selected than the ones in ProsAudit, which compensates this by having much more stimuli. Still, in the protosyntax task, humans perform better than the models, acting as a topline.  On the lexical task, both models score higher than humans, who were less confident in their ratings than in the protosyntax task. This may be due to the wording of the experiment, asking the participants to choose the most natural sentence without any cues to focus on the lexical aspect. Moreover, in this task, both conditions sound pretty unnatural, as the ``correct" stimuli have a pause placed at a word boundary with no prosodic break. This hypothesis is supported by additional analyses indicating lower participant confidence in the lexical task compared to the protosyntax task.

\begin{table}[]
\renewcommand{\arraystretch}{0.9}
\caption{ProsAudit accuracy scores (\%) for the human evaluation and best performing STELA and GSLM models, on the test and human subset.}
\label{tab:human_comparison}
\centering
\scalebox{0.9}{
\begin{tabular}{@{}lllll@{}}
\toprule
                     & \multicolumn{2}{l}{protosyntax} & \multicolumn{2}{l}{lexical} \\
                     & test           & human set   & test         & human set \\ \midrule
humans native        & -    & 80.50           & -  & 60.38         \\
STELA    &    74.86       & 73.32           &  68.28       & 71.18         \\
pGSLM &  65.94         & 64.88           &  72.88       & 73.33        \\ \bottomrule
\end{tabular}
}
\vspace{-5mm}
\end{table}

\vspace{0.2cm}\noindent\textbf{Effect of size.} Results averaged over all STELA models from a same size of the training dataset, both in English and French, are presented in Figure \ref{fig:size_effect}. Focusing solely on the English models (solid line), we can note multiple things. First, even models trained on a small amount of data (50 hours) achieve above-chance performance in the prosodic tasks, indicating that structural prosodic knowledge can be acquired from limited data. Additionally, the results show a clear trend of improvement in performance as the size of the training dataset increases, particularly for the lexical task, which exhibits a logarithmic growth. However, in the case of the protosyntax task, the study observes a plateau in the improvement of performance between 1,600 and 3,200 hours, potentially indicating a ceiling effect, meaning that beyond this point, increasing the size of the dataset does not have a significant impact on the model's performance.

\vspace{0.2cm}\noindent\textbf{Effect of nativeness.} We analysed the performance of models trained on French data when evaluated on English prosodic tasks, as depicted by the dotted line in Figure \ref{fig:size_effect}. When considering only models trained on the largest amount of data, it is found that there is a clear native advantage for English models, achieving better performance than their French counterparts in both tasks. This difference is more pronounced in the lexical task, where a strong lexical knowledge of English is a key factor for success. A comparison of the overall developmental curves reveals that while the performance of French models improves in the protosyntax task with increasing training data, this is not the case for the lexical task. The lack of improvement for French models in this second task results in a widening performance gap between native and non-native models. These findings, although unsurprising, indicate that non-native models cannot acquire the necessary lexical knowledge to excel in this task solely through additional training data.

\section{Discussion \& Conclusion}

We introduced ProsAudit, a zero-shot benchmark for measuring the English prosodic knowledge of speech SSL models, made of two subtasks. The protosyntax task assesses the model's understanding of structural prosody at the syntactic level. All the models we evaluated performed well above chance in this task, indicating that this knowledge is embedded in speech SSL models. Besides, models trained on another language perform relatively well in this task, which is in line with findings in humans \cite{ludusan2021effect}. This suggests that some prosodic knowledge can be universal \cite{endress2010word}. However, the gap between native and non-native models increases with more data, suggesting that some prosodic cues are language-specific, indicating that the models are in line with findings from the psycholinguistics literature \cite{hohle2009language}. 

The lexical task requires both hierarchical prosodic knowledge at the sentence level and some lexical knowledge, as prosody is not always enough to differentiate between word boundary and within word breaks. The GSLM models (with and without prosody) performed better on this task than the protosyntax one.  This could be because they have some strong lexical knowledge of English (these models perform relatively well on lexical metrics, see \cite{lakhotia2021generative, dunbar2021zero}), levelling up with their knowledge of prosody. Conversely, the STELA models do not perform as well on this task as the protosyntax one, although they still score way above chance, suggesting their lexical knowledge is not strong enough to surpass the protosyntax task. Unsurprisingly, there is a much larger native effect in this task, with non-native models performing only slightly above chance, regardless of the size of the training set. In contrast, the native models' performance is strongly correlated with data size. Further research could examine how much lexical knowledge in a model correlates with their performance on the ProsAudit lexical task.

\ifinterspeechfinal \vspace{-0.7em} \else  \fi

We also found that the pGSLM models, despite supposedly encoding prosody-specific features and yielding better mean opinion scores on downstream tasks \cite{kharitonov2022text}, score only slightly better on the protosyntax and lexical tasks than their vanilla GSLM counterparts. Two things could explain these results. First, we only compute the pseudo-probability taking into account the loss at the token level (discarding losses at the pitch and duration levels), and finding news ways to generate this pseudo-probability by taking into account these two components could increase the models' scores. Second, the benchmark we propose here only evaluates specific aspects of prosody at the structural level, while other aspects of prosody as emotion, not evaluated in this benchmark, could be more related to the metrics presented in \cite{kharitonov2022text}.

\ifinterspeechfinal \vspace{-0.7em} \else  \fi

To conclude, we propose for the first time a new zero-shot benchmark for evaluating structural prosodic knowledge in speech models, along with a human evaluation topline. We hope that it will inspire further research to enhance the prosodic capabilities of SSL models. This is crucial, as research has demonstrated that models with better prosodic understanding lead to more advanced generative speech models (as seen in \cite{kharitonov2022text}). 
\ifinterspeechfinal 
Additionally, the benchmark is now incorporated into Track 4 of the Zero-Resource Speech challenge \cite{nguyen2020zero}, providing a platform for continuous improvement and comparison of models.
\else    
Additionally, the benchmark is now made available online, providing a platform for continuous improvement and comparison of models.
\fi
In the future, we aim to develop this benchmark in languages other than English, and to expand the benchmark to subtypes targeting other aspects of prosody.

\ifinterspeechfinal 

 \vspace{-0.7em}

\noindent \footnotesize \textbf{Acknowledgments.}
MS's work was partly funded by l'Agence de l'Innovation de Défense and performed using HPC resources from GENCI-IDRIS (Grant 20XX-AD011012315). ED in his EHESS role was supported in part by the Agence Nationale pour la Recherche (ANR-17-EURE-0017 Frontcog, ANR-10-IDEX-0001-02 PSL*, ANR-19-P3IA-0001 PRAIRIE 3IA Institute) and a grant from CIFAR (Learning in Machines and Brains).
\else

\fi

\bibliographystyle{IEEEtran}
\bibliography{mybib}

\begin{thebibliography}{10}
\providecommand{\url}[1]{#1}
\csname url@samestyle\endcsname
\providecommand{\newblock}{\relax}
\providecommand{\bibinfo}[2]{#2}
\providecommand{\BIBentrySTDinterwordspacing}{\spaceskip=0pt\relax}
\providecommand{\BIBentryALTinterwordstretchfactor}{4}
\providecommand{\BIBentryALTinterwordspacing}{\spaceskip=\fontdimen2\font plus
\BIBentryALTinterwordstretchfactor\fontdimen3\font minus
  \fontdimen4\font\relax}
\providecommand{\BIBforeignlanguage}[2]{{%
\expandafter\ifx\csname l@#1\endcsname\relax
\typeout{** WARNING: IEEEtran.bst: No hyphenation pattern has been}%
\typeout{** loaded for the language `#1'. Using the pattern for}%
\typeout{** the default language instead.}%
\else
\language=\csname l@#1\endcsname
\fi
#2}}
\providecommand{\BIBdecl}{\relax}
\BIBdecl

\bibitem{baevski2020wav2vec}
A.~Baevski, Y.~Zhou, A.~Mohamed, and M.~Auli, ``wav2vec 2.0: A framework for
  self-supervised learning of speech representations,'' \emph{Advances in
  neural information processing systems}, vol.~33, pp. 12\,449--12\,460, 2020.

\bibitem{oord2018representation}
A.~v.~d. Oord, Y.~Li, and O.~Vinyals, ``Representation learning with
  contrastive predictive coding,'' \emph{arXiv preprint arXiv:1807.03748},
  2018.

\bibitem{riviere2020unsupervised}
M.~Riviere, A.~Joulin, P.-E. Mazar{\'e}, and E.~Dupoux, ``Unsupervised
  pretraining transfers well across languages,'' in \emph{ICASSP 2020-2020 IEEE
  International Conference on Acoustics, Speech and Signal Processing
  (ICASSP)}.\hskip 1em plus 0.5em minus 0.4em\relax IEEE, 2020, pp. 7414--7418.

\bibitem{hsu2021hubert}
W.-N. Hsu, B.~Bolte, Y.-H.~H. Tsai, K.~Lakhotia, R.~Salakhutdinov, and
  A.~Mohamed, ``Hubert: Self-supervised speech representation learning by
  masked prediction of hidden units,'' \emph{IEEE/ACM Transactions on Audio,
  Speech, and Language Processing}, vol.~29, pp. 3451--3460, 2021.

\bibitem{dunbar2017zero}
E.~Dunbar, X.~N. Cao, J.~Benjumea, J.~Karadayi, M.~Bernard, L.~Besacier,
  X.~Anguera, and E.~Dupoux, ``The zero resource speech challenge 2017,'' in
  \emph{2017 IEEE Automatic Speech Recognition and Understanding Workshop
  (ASRU)}.\hskip 1em plus 0.5em minus 0.4em\relax IEEE, 2017, pp. 323--330.

\bibitem{nguyen2020zero}
T.~A. Nguyen, M.~de~Seyssel, P.~Roz{\'e}, M.~Rivi{\`e}re, E.~Kharitonov,
  A.~Baevski, E.~Dunbar, and E.~Dupoux, ``The zero resource speech benchmark
  2021: Metrics and baselines for unsupervised spoken language modeling,'' in
  \emph{NeuRIPS Workshop on Self-Supervised Learning for Speech and Audio
  Processing}, 2020.

\bibitem{dunbar2021zero}
E.~Dunbar, M.~Bernard, N.~Hamilakis, T.~A. Nguyen, M.~de~Seyssel, P.~Rozé,
  M.~Rivière, E.~Kharitonov, and E.~Dupoux, ``{The Zero Resource Speech
  Challenge 2021: Spoken Language Modelling},'' in \emph{Proc. Interspeech
  2021}, 2021, pp. 1574--1578.

\bibitem{yang2021superb}
S.~wen Yang, P.-H. Chi, Y.-S. Chuang, C.-I.~J. Lai, K.~Lakhotia, Y.~Y. Lin,
  A.~T. Liu, J.~Shi, X.~Chang, G.-T. Lin, T.-H. Huang, W.-C. Tseng, K.~tik Lee,
  D.-R. Liu, Z.~Huang, S.~Dong, S.-W. Li, S.~Watanabe, A.~Mohamed, and
  H.~yi~Lee, ``{SUPERB: Speech Processing Universal PERformance Benchmark},''
  in \emph{Proc. Interspeech 2021}, 2021, pp. 1194--1198.

\bibitem{cutler1997prosody}
A.~Cutler, D.~Dahan, and W.~Van~Donselaar, ``Prosody in the comprehension of
  spoken language: A literature review,'' \emph{Language and speech}, vol.~40,
  no.~2, pp. 141--201, 1997.

\bibitem{dahan2015prosody}
D.~Dahan, ``Prosody and language comprehension,'' \emph{Wiley Interdisciplinary
  Reviews: Cognitive Science}, vol.~6, no.~5, pp. 441--452, 2015.

\bibitem{kharitonov2022text}
E.~Kharitonov, A.~Lee, A.~Polyak, Y.~Adi, J.~Copet, K.~Lakhotia, T.~A. Nguyen,
  M.~Riviere, A.~Mohamed, E.~Dupoux \emph{et~al.}, ``Text-free prosody-aware
  generative spoken language modeling,'' in \emph{Proceedings of the 60th
  Annual Meeting of the Association for Computational Linguistics (Volume 1:
  Long Papers)}, 2022, pp. 8666--8681.

\bibitem{lin2023utility}
G.-T. Lin, C.-L. Feng, W.-P. Huang, Y.~Tseng, T.-H. Lin, C.-A. Li, H.-y. Lee,
  and N.~G. Ward, ``On the utility of self-supervised models for
  prosody-related tasks,'' in \emph{2022 IEEE Spoken Language Technology
  Workshop (SLT)}.\hskip 1em plus 0.5em minus 0.4em\relax IEEE, 2023, pp.
  1104--1111.

\bibitem{ludusan2021effect}
B.~Ludusan, M.~Morii, Y.~Minagawa, and E.~Dupoux, ``The effect of different
  information sources on prosodic boundary perception,'' \emph{JASA Express
  Letters}, vol.~1, no.~11, p. 115203, 2021.

\bibitem{ostendorf1995boston}
M.~Ostendorf, P.~J. Price, and S.~Shattuck-Hufnagel, ``The boston university
  radio news corpus,'' \emph{Linguistic Data Consortium}, pp. 1--19, 1995.

\bibitem{silverman1992tobi}
K.~E. Silverman, M.~E. Beckman, J.~F. Pitrelli, M.~Ostendorf, C.~W. Wightman,
  P.~Price, J.~B. Pierrehumbert, and J.~Hirschberg, ``Tobi: A standard for
  labeling english prosody.'' in \emph{ICSLP}, vol.~2, 1992, pp. 867--870.

\bibitem{lakhotia2021generative}
K.~Lakhotia, E.~Kharitonov, W.-N. Hsu, Y.~Adi, A.~Polyak, B.~Bolte, T.-A.
  Nguyen, J.~Copet, A.~Baevski, A.~Mohamed \emph{et~al.}, ``On generative
  spoken language modeling from raw audio,'' \emph{Transactions of the
  Association for Computational Linguistics}, vol.~9, pp. 1336--1354, 2021.

\bibitem{kahn2020libri}
J.~Kahn, M.~Riviere, W.~Zheng, E.~Kharitonov, Q.~Xu, P.-E. Mazar{\'e},
  J.~Karadayi, V.~Liptchinsky, R.~Collobert, C.~Fuegen \emph{et~al.},
  ``Libri-light: A benchmark for asr with limited or no supervision,'' in
  \emph{ICASSP 2020-2020 IEEE International Conference on Acoustics, Speech and
  Signal Processing (ICASSP)}.\hskip 1em plus 0.5em minus 0.4em\relax IEEE,
  2020, pp. 7669--7673.

\bibitem{lavechin2022statistical}
M.~Lavechin, M.~de~Seyssel, H.~Titeux, H.~Bredin, G.~Wisniewski, A.~Cristia,
  and E.~Dupoux, ``Can statistical learning bootstrap early language
  acquisition? a modeling investigation,'' \emph{PsyArXiv preprint
  PsyArXiv:rx94d}, 2022.

\bibitem{endress2010word}
A.~D. Endress and M.~D. Hauser, ``Word segmentation with universal prosodic
  cues,'' \emph{Cognitive psychology}, vol.~61, no.~2, pp. 177--199, 2010.

\bibitem{hohle2009language}
B.~H{\"o}hle, R.~Bijeljac-Babic, B.~Herold, J.~Weissenborn, and T.~Nazzi,
  ``Language specific prosodic preferences during the first half year of life:
  Evidence from german and french infants,'' \emph{Infant Behavior and
  Development}, vol.~32, no.~3, pp. 262--274, 2009.

\end{thebibliography}

\end{document}